\documentclass[letterpaper, 10 pt, conference]{ieeeconf}  

\IEEEoverridecommandlockouts                              

\usepackage[caption=false, font=footnotesize]{subfig}
\usepackage{graphicx} 
\usepackage{algorithm} 
\usepackage{algpseudocode} 
\usepackage{amsmath} 
\usepackage{amssymb}  
\usepackage{siunitx}
\usepackage{wrapfig}
\usepackage{array}
\usepackage{float}
\usepackage{hyperref}
\usepackage{booktabs}
\title{\LARGE \bf
SoftHand Model-W: A 3D-Printed, Anthropomorphic, Underactuated Robot Hand with Integrated Wrist and Carpal Tunnel
}

\author{Dhillon B. Merritt, Christopher J. Ford, Haoran Li, Malia Smith, Zhixing Chen, \\Efi Psomopoulou$^*$, Nathan F. Lepora$^*$,  
\thanks{All authors are with the School of Engineering Mathematics and Technology, and Bristol Robotics Laboratory, University of Bristol, Bristol, BS8 1UB, U.K. * Equal contribution. Correspondance to n.lepora@bristol.ac.uk}
\thanks{This research was partly funded by the Advanced Research + Invention Agency (ARIA) under grant ``Democratising Co-Design of Hardware and Control for Robot Dexterity''}
}

\begin{document}

\maketitle
\thispagestyle{empty}
\pagestyle{empty}

\begin{abstract}
This paper presents the SoftHand Model-W: a 3D-printed, underactuated, anthropomorphic robot hand  based on the Pisa/IIT SoftHand, with an integrated antagonistic tendon mechanism and 2 degree-of-freedom tendon-driven wrist. These four degrees-of-acuation provide active flexion and extension to the five fingers, and active flexion/extension and radial/ulnar deviation of the palm through the wrist, while preserving the synergistic and self-adaptive features of such SoftHands. A carpal tunnel-inspired tendon routing allows remote motor placement in the forearm, reducing distal inertia and maintaining a compact form factor. The SoftHand-W is mounted on a 6-axis robot arm and tested with two reorientation tasks requiring coordination between the hand and arm's pose: cube stacking and in-plane disc rotation. Results comparing task time, arm joint travel, and configuration changes with and without wrist actuation show that adding the wrist reduces compensatory and reconfiguration movements of the arm for a quicker task-completion time. Moreover, the wrist enables pick-and-place operations that would be impossible otherwise. Overall, the SoftHand Model-W demonstrates how proximal degrees of freedom are key to achieving versatile, human-like manipulation in real world robotic applications, with a compact design enabling deployment in research and assistive settings.
\end{abstract}


\section{INTRODUCTION}

Dexterous manipulation remains a key challenge in robotics, limiting the range of tasks that robots can perform in unstructured environments \cite{billard2019trends}. Achieving a level of dexterity equivalent to that of humans would allow robots to interact with a diverse range of objects, perform fine motor tasks, and adapt to unforeseen situations in industrial, service, and assistive applications \cite{okamura2000overview,bicchi2000robotic}. One of the most promising pathways to this goal is anthropomorphic design, since much of our infrastructure, tools, and technology are tailored to the dimensions and motion capabilities of the human hand \cite{ritter2015hands}. By replicating the structure and actuation principles of the human hand, robotic systems could better integrate into human environments, leveraging the same manipulation strategies that we use naturally.

\begin{figure}[!t]
    \centering
    \includegraphics[width=0.48\textwidth]{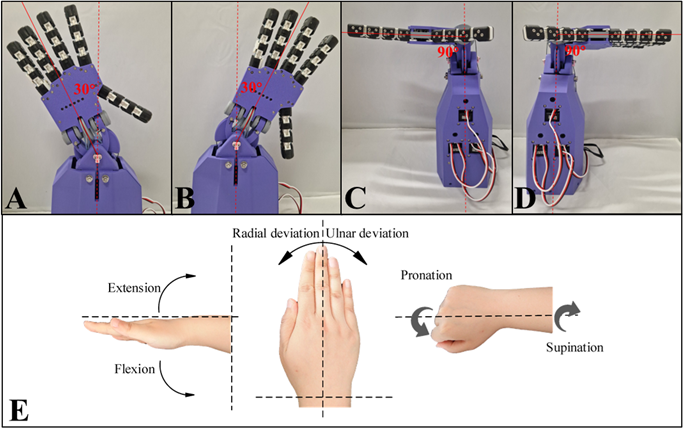}
    \caption{Range of motion of the SoftHand Model-W in the ulnar (A) and radial (B) deviations, and in flexion (C) and extension (D). Corresponding degrees of freedom of the human wrist (E) are shown below (based on \cite{huang2017wearable}).}
    \label{fig:main_photo}
    \vspace{-1em}
\end{figure}

Although there have been major advances in anthropomorphic robot hand design~\cite{piazza2019century}, a significant gap remains in replicating the full spectrum of human manipulation capabilities. The wrist plays a significant role in our dexterity (Fig.~\ref{fig:main_photo}E), contributing 3 additional Degrees of Freedom (DoF) \cite{fan2022prosthetic,montagnani2015finger}. These expand the range of reachable orientations and enable efficient in-hand manipulation \cite{fan2022prosthetic,bajaj2019state}. Despite this, research into robot hands has prioritised increasing the number of actuated joints in the fingers whilst leaving the wrist under-represented or entirely fixed \cite{capsi2023softhand,jonna2025hadar,gao2021dexterous}. As a result, when mounted on robot arms, these hands will require large compensatory arm motions to achieve the desired hand pose, which is inefficient and impractical in confined spaces.

This paper proposes a novel 3D-printed, anthropomorphic, tendon-driven robotic hand based on an evolved Pisa/IIT softHand design~\cite{catalano2014adaptive} with an integrated 2-DoF wrist (Fig.~\ref{fig:main_photo}A-D). The original Pisa/IIT SoftHand has just one motor to close the hand, with a passive reset using elastic ligaments. More recently, the BPI SoftHand-A~\cite{li2024tactile} extended this functionality with a second motor and agonist/antagonist tendons on each finger, preserving the synergistic action of the SoftHand while greatly improving grip force, reactivity and control. Here we explore how this can be integrated with degrees of freedom in a wrist, which raises challenges on how to preserve the antagonistic and synergistic action while coordinating the wrist and hand. To do this, we introduce a biomimetic version of the human carpal tunnel, with the finger tendons routed through central channels in the wrist. This allows for remote motor placement, keeping the distal structure lightweight and preserving a compact form factor. Additionally, it decouples finger and wrist actuation, enabling a wider range of poses. We then consider how the addition of a wrist improves the performance on manipulation tasks, with a focus on object handling and reorientation.


Our main contributions are:\\
\noindent 1) {\em 3D-printed, human-scale, tendon-driven, anthropomorphic SoftHand:} The design of the (bulky) BRL SoftHand~\cite{li2022brl,li2024tactile} is miniaturized to create an accessible, low-cost robotic hand that features underactuated, tendon-driven fingers and matches the form of the human hand.\\
\noindent 2) {\em Integrated 2-DoF wrist with carpel tunnel tendon routing:} The design includes a 2-DoF wrist that enables flexion/extension and radial/ulnar deviation. A carpal tunnel routes the finger tendons through the wrist, decoupling the fingers from the wrist and allowing remote motor placement.\\
\noindent 3) {\em Validation with object manipulation:} Two manipulation tasks are introduced to demonstrate the utility of the wrist: in-hand object rotation of a disk through 90\unit{\degree} and stacking six cubes from differing initial orientations and positions. \\
We refer to the presented design as the \textit{SoftHand Model-W}, building on the lineage of the Pisa/IIT and BRL SoftHand designs, with `W' denoting the addition of an active wrist.

\section{BACKGROUND AND RELATED WORK}\label{Background}
The Pisa/IIT SoftHand \cite{catalano2014adaptive,della2018toward} employs underactuated, tendon-driven mechanics and the concept of ``adaptive synergies" to produce robust, versatile grasps. Adaptive synergies refer to coordinated patterns of joint movement that are derived from human hand usage, allowing a single actuator to drive multiple degrees of freedom in a way that conforms naturally to objects being grasped. The Pisa/IIT SoftHand uses only one actuator for the finger flexor tendons, with extension achieved using passive elastic components. Its compliant, underactuated joints and adaptive synergies allow passive adaptation to object shapes, greatly simplifying control without sacrificing dexterity, particularly with tactile sensing~\cite{ford2023tactile, ford2025shear}. Whilst the SoftHand excels in its simplicity and grasping capabilities, it lacks a wrist, limiting its efficiency in many tasks. Capsi \textit{et al.} \cite{capsi2023softhand} did extend the Pisa/IIT SoftHand design to include a {\em passive} wrist for a transradial prosthetic, capable of pronation and supination. However, they found that the lack of an {\em active} wrist necessitated large compensatory movements which were uncomfortable for the user to perform and could compromise EMG sensor contact with the skin \cite{capsi2023softhand}.

Li \textit{et al.} later introduced two derivatives of the SoftHand. First, the BRL/Pisa/IIT SoftHand \cite{li2022brl} that preserved the adaptive synergy concept but adopted a mostly 3D-printed construction to reduce cost and ease customization and manufacturability. This made hand prototyping more accessible but increased its size compared to the original SoftHand that was the size of a human hand. The more recent Tactile SoftHand-A \cite{li2024tactile} incorporated antagonistic tendon pairs for the fingers with active flexion and extension, which enables controlled stiffness and greater grasp strength and responsiveness as the finger motion no longer needs to overcome passive elastic extension forces. Whilst these designs expand functionality and accessibility, they still lack a wrist and the increase in size reduces anthropomorphism.

Robotic wrists vary widely in architecture and actuation, with designs ranging from simple passive joints \cite{capsi2023softhand} to complex multi-DoF systems \cite{kim2018quaternion}, but all fall under two main categories: serial and parallel wrists. Serial wrists consist of one or more joints arranged in a chain, allowing for simple motion analysis and large workspaces for a given DoF count. These can be directly driven or actuated remotely using tendons or belts \cite{fan2022prosthetic}. However, they are at risk of accumulative errors and deformation as heavy loads are transferred through the entirety of the wrist structure. Parallel wrists, on the other hand, use multiple kinematic chains between the base and end-effector, providing higher stiffness, load capacity, and precision \cite{peticco2025dexwrist}. This comes at the cost of a reduced workspace to compensate for the increased complexity of singularity avoidance and control \cite{fan2022prosthetic}. 

Due to the 3D-printed construction and tendon-driven actuation of the considered SoftHand, a serial wrist configuration was chosen. This allows for both the hand and wrist to be actuated using tendons and the same type of servos, reducing the complexity of the control system. This facilitates the inclusion of a carpal tunnel, enabling remote motor placement for reduced distal inertia. Altogether this results in a SoftHand design that is suited for dexterous manipulation, which will be demonstrated later in this paper.

\section{RESEARCH METHODOLOGY}\label{Methodology}
\subsection{Implementation}\label{sec:implementation}
\subsubsection{Overall Construction}

\begin{figure}[t]
    \centering
    \includegraphics[width=0.48\textwidth]{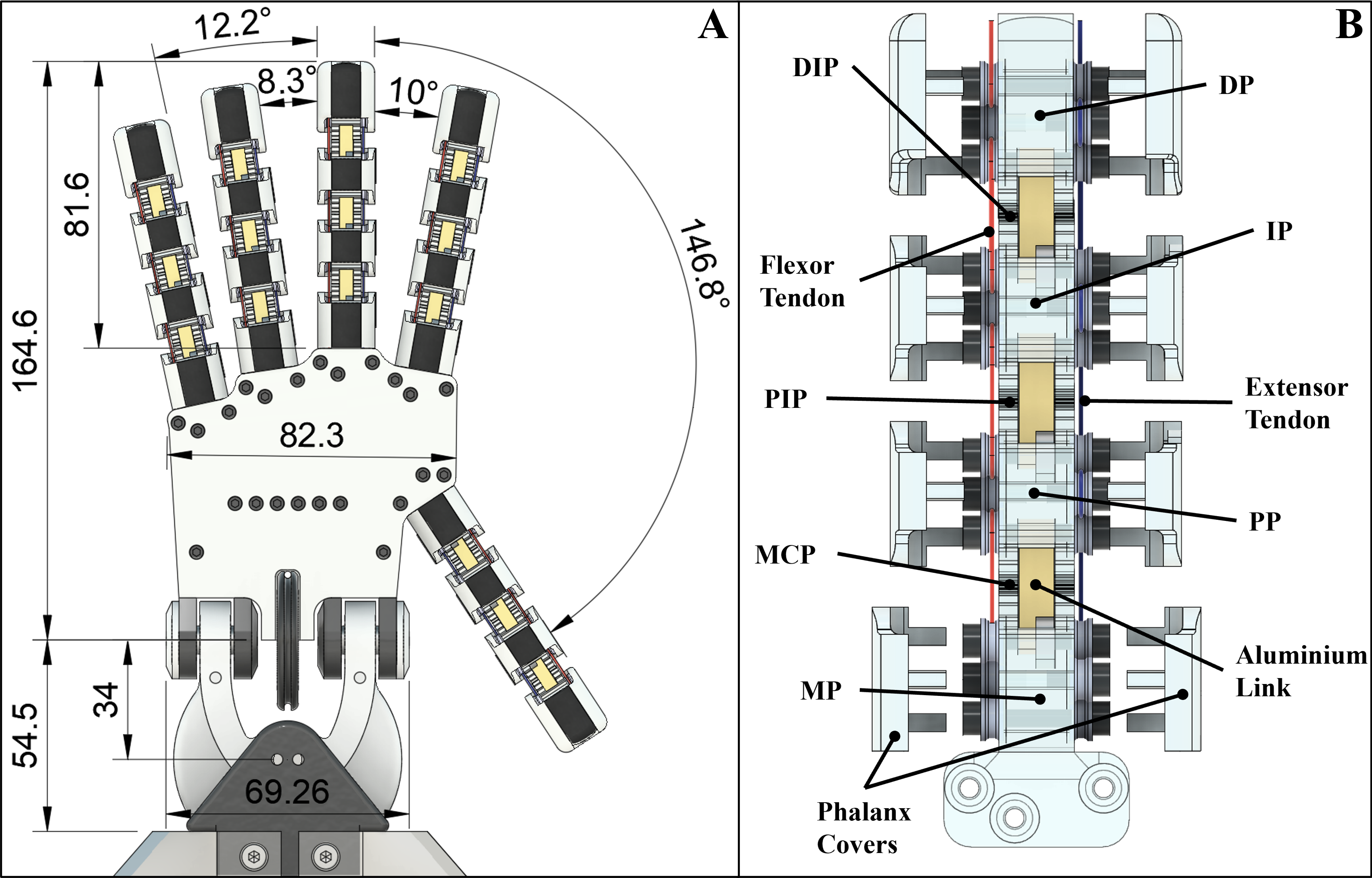}
    \caption{Overview of the SoftHand-W. A) Technical drawing showing the palmer side with finger, palm, and write dimensions in \unit{mm}. B) CAD model of a finger (palmer side up), with flexor (red) and extensor (blue) tendons.}
    \label{fig:overview}
    \vspace{-1em}
\end{figure}

The hand and fingers are based on the antagonistic SoftHand design by Li \textit{et al.} \cite{li2024tactile}. Excluding the wrist, the new SoftHand Model-W is constructed from two main components: the palm and the fingers, with a total of 15 DoFs (Fig.~\ref{fig:overview}). Overall, the hand is 164.6\unit{mm} tall when measuring from the base of the palm to the middle fingertip, with all fingers identical with a length of 81.6\,\unit{mm}. In comparison, the 50th percentile adult hand length is 189\,\unit{mm} (male) and 172\,\unit{mm} (female), whilst the 5th-95th percentile spans 173-205\,\unit{mm} (male) and 159-189\,\unit{mm} (female) \cite{pheasant2018bodyspace}. The SoftHand-W therefore falls within a reasonable range of dimensions, supporting ergonomic interaction with common handles, tools, and objects sized for human users. Due to the lack of finger splay functionality, the hand is designed with fixed finger angles of -146.8\unit{\degree} (thumb), -10\unit{\degree} (index), 0\unit{\degree} (middle), 8\unit{\degree} (ring), and 12.2\unit{\degree} (little), from the vertical.

The majority of the hand is 3D-printed in PLA using a standard FDM printer, with the exception of some components that are printed in resin on an SLA printer for improved dimensional accuracy and strength. Additionally, M2 screws, bearings, PTFE tubing, nylon thread, servos, and basic CNC-milled aluminium parts complete the assembly.

\subsubsection{Fingers}
Each finger is comprised of four PLA phalanges. Like the human hand, these comprise the distal phalanx (DP), intermediate phalanx (IP), and proximal phalanx (PP). In addition to these is the metacarpal phalanx (MP), which is fixed and connects the finger to the palm. These are connected using aluminium links and gear engagement to form the distal interphalangeal (DIP), proximal interphalangeal (PIP), and metacarpophalangeal (MCP) joints (Fig.~\ref{fig:overview}), corresponding to each finger's 3 DoFs. 

Each finger is actuated using nylon flexor and extensor tendons, both of which start in the fingertip. Each phalanx has three U-groove bearings (5\,mm dia.) on each side fastened with M2$\times$8 screws, which the tendons are routed through, as shown in Fig.~\ref{fig:overview}. On the right side of the phalanx, these bearings are arranged in an equilateral triangle, orientated such that linear displacement of the flexor tendon results in rotation of the phalanx and flexion of the finger joints. The extensor tendon is routed similarly on the left side of the phalanx, but the bearings are inverted enabling extension of the finger joints. Covers are slotted onto each phalanx on either side, protecting the bearings and preventing the tendons from slipping off. These were printed in resin on an SLA printer as the small pegs that hold them in place were prone to snapping when printed on an FDM printer.

\begin{figure}[t]
    \centering
    \includegraphics[width=0.48\textwidth]{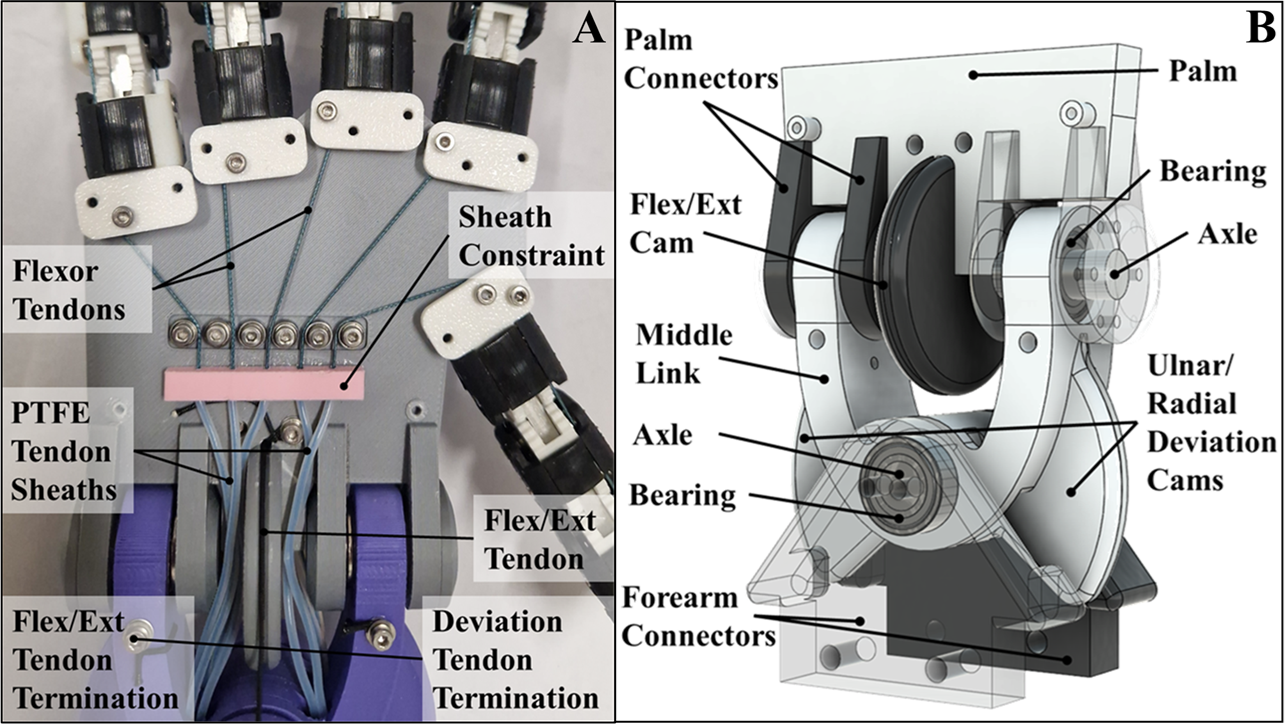}
    \caption{Wrist mechanism and tendon routing. A) Tendon routing through the palm and into the wrist (routing on the back of the hand is similar). B) CAD model of the 2-DoF wrist mechanism, showing the internal bearings and axles as well as palm and forearm connectors.}
    \label{fig:wrist_mechanism}
    \vspace{-1em}
\end{figure}

\subsubsection{Tendon Routing and Carpal Tunnel}
After exiting the fingers, the tendons are routed through a series of pulleys to the centre of the palm. They then enter the wrist through a carpal tunnel, which allows for remote motor placement in the forearm (Fig.~\ref{fig:wrist_mechanism}). It is important that the design mitigates undesired consequence of coupling wrist and finger motion, with extension of the wrist inducing flexion in the fingers, and vice versa. This issue is addressed by the use of PTFE `tendon sheaths' through the carpal tunnel. Constrained at each end, these work like Bowden cables, allowing the tendons to slide freely through the wrist with minimal friction and changing the reference point for the tendons' movement. Instead of the tendons moving relative to output and input components (the fingers and servos), they now move relative to the stationary sheaths, which are capable of bending with the wrist without changing in length.

\subsubsection{Wrist}\label{wrist}
The wrist uses a serial rotational-rotational (RR) configuration to achieve pitch and yaw (flexion/extension and ulnar/radial deviation in humans) as shown in Fig.~\ref{fig:wrist_mechanism}. The third DoF of the human wrist, pronation/supination (Fig.~\ref{fig:main_photo}), is not included in this design as most robot arms already terminate with this DoF to rotate the end effector~\cite{villani2018survey}. Additionally, in humans, the whole forearm is used to achieve this motion, with the two bones of the forearm (the radius and ulnar) rotating around each other. It therefore makes sense to keep this DoF in the arm rather than the wrist. This also helps to keep the wrist compact, with a height, width, and depth of 54.5\,\unit{mm}, 69.3\,\unit{mm}, and 41\,\unit{mm} respectively (Fig.~\ref{fig:overview}).

Each joint uses an 8\,\unit{mm}-long aluminium bar as an axle, and 16\unit{mm}OD 8\unit{mm}ID bearings for smoother operation, reduce lateral play, and strengthen the joint. The two rotational joints are orthogonal to each other, with a small offset of 34\,\unit{mm} between the two axes and a 48\,\unit{mm} offset between the flexion axis and the centre of the palm. Using distal Denavit-Hartenberg parameters (joint $i$: $a_i$, $\alpha_i$, $d_i$, $\theta_i$) \cite{corke2007simple}, the ulnar/radial deviation joint ($i=1$) and flexion/extension joint ($i=2$) are modeled as shown in Table \ref{tab:dh}. Here $a_1$ is the offset between wrist DoFs and $a_2$ is the offset between the flexion axis to the centre of the palm.

\begin{table}[H]
    \vspace{-.5em}
    \centering
    \begin{tabular}{c|cccc}
    joint, $i$   & $a_i$ (\unit{mm}) & $\alpha_i$ (\unit{rad})    & $d_i$ (\unit{mm})   & $\theta_i$ (\unit{rad})\\
    \hline
    1       & 34      & $\pi/2$       & 0   & $\theta_1$\\
    2       & 48      & 0               & 0         & $\theta_2$\\
    \end{tabular}
    \caption{Distal Denavit-Hartenberg parameters of the wrist. }
    \vspace{-2em}
    \label{tab:dh}
\end{table}

Like the fingers, these joints are tendon-driven, with integrated cams used to convert the linear motion of the tendons into rotational motion at the joints consistently, throughout the range of motion. These have a constant radius, as do the tendon spools (Fig.~\ref{fig:actuation_box}), resulting in a linear relationship between the servo position and wrist position. A U-shaped channel is used to prevent the tendons from slipping off the cams when the wrist is in extreme positions.

\subsubsection{Actuation and Control}
Both the finger and wrist tendons are terminated on spools in the `forearm'. The flexor tendons are all terminated on one spool, and the extensor tendons on another. This allows for antagonistic actuation of the fingers, with the flexor tendons pulling the fingers into a closed position, and the extensor tendons pulling them into an open position. The wrist tendons are similarly terminated on two spools, one for each DoF.

\begin{figure}[t]
    \centering
    \includegraphics[width=0.8\columnwidth]{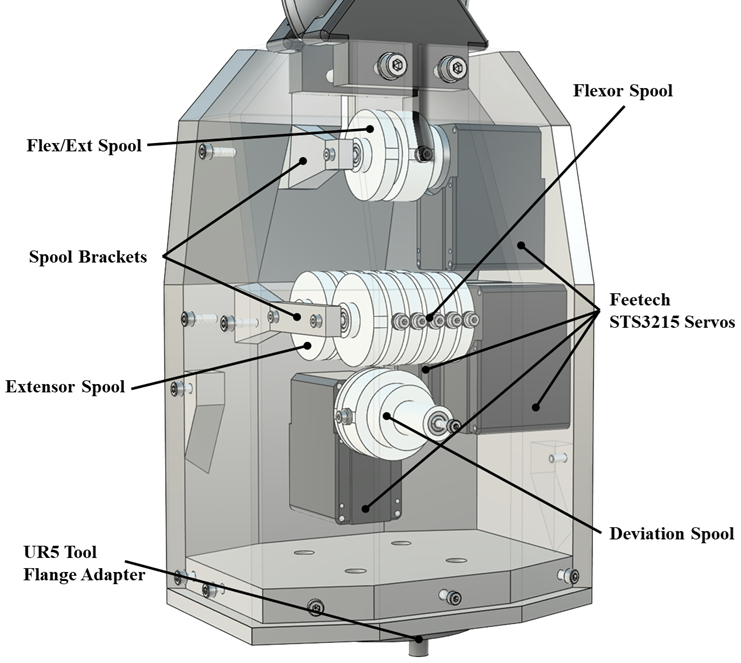}
    \caption{CAD model of the `forearm' - an actuation box containing four Feetech STS3215 servos and tendon spools, which enable finger flexion, finger extension, radial/ulnar deviation, and flexion/extension of the wrist. Mounting for the UR5 arm is achieved via a tool flange adapter at the base.}
    \label{fig:actuation_box}
    \vspace{-1em}
\end{figure}

The BPI SoftHand designs by Li \textit{et al.} require that the tendons be terminated with a knot on either end, both at the spool and the fingertip \cite{li2022brl}, \cite{li2024tactile}. However, this can make tensioning the tendons difficult, which is of particular concern for the finger tendons. Since the flexor tendons are all terminated on the same spool, they must all be tensioned equally to ensure that the fingers flex uniformly. The same is true for the extensor tendons. Instead, an M2 bolt with two washers was used for tendon termination, as shown in Figs.~\ref{fig:wrist_mechanism},~\ref{fig:actuation_box}. This allows for easy tensioning of the tendons by wrapping the tendon around the bolt, before clamping it between the two washers by tightening the bolt.

The spools are actuated using four Feetech STS3215 servos, two for the fingers and two for the wrist. These were chosen for their low cost, high torque output (19.5\,\unit{kg.cm} at 7.4\unit{V}), support of daisy chaining, and ease of control. They are mounted in the forearm with a tendon spool on each (Fig.~\ref{fig:actuation_box}). Brackets are used to prevent the spools from deflecting laterally, ensuring maximum transfer of force to the tendons. The servos are controlled using Feetech's URT-1 controller for simple serial communication with the servos. The controller is connected to a PC via USB, and the servos are controlled with Python. This method of control was chosen compatibility with the control of the UR5 arm upon which the hand and wrist will be mounted for the manipulation tasks (see Section \ref{sec:experiment_method}). 


Finger control uses open and close commands based on calibrated position values, but wrist control is more complicated. As mentioned in Section \ref{wrist}, the cams used in the wrist joints have a constant radius, as do the tendon spools. This results in a linear relationship between the servo position and wrist position. The wrist is therefore controlled using an open-loop proportional controller, with the servo position set to the desired wrist position multiplied by a constant gain. This controller converts requested angles into the servo position around calibrated centre values. With the 2 additional DoFs in the wrist, the tool centre point (TCP) is no longer constant when the SoftHand is mounted on the arm. To account for this, the two wrist angles and their offsets are used to compute the new tool position via the forward kinematics described above (Section \ref{wrist}).

\begin{figure}[t]
    \centering
  \subfloat[CAD model of the 3D-printed rotation test stimulus. The hand is tasked with grasping the disc from above and rotating it 90\unit{\degree} anti-clockwise.]{%
       \includegraphics[width=0.48\linewidth]{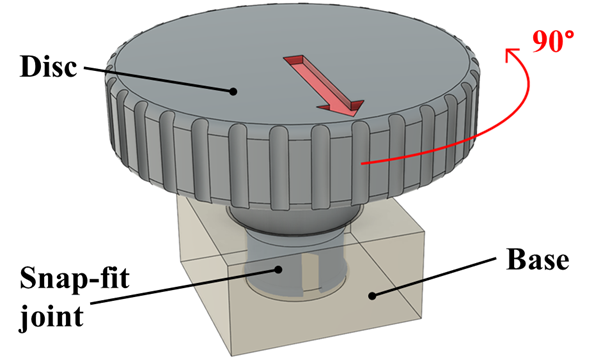}}
       \label{fig:disk}
    \hfill
  \subfloat[Stacking test results: A) without actuation of the wrist; B) With actuation of the wrist. Cubes are numbered in the order that they are picked and placed. \label{fig:stack}]{%
        \includegraphics[width=0.48\linewidth]{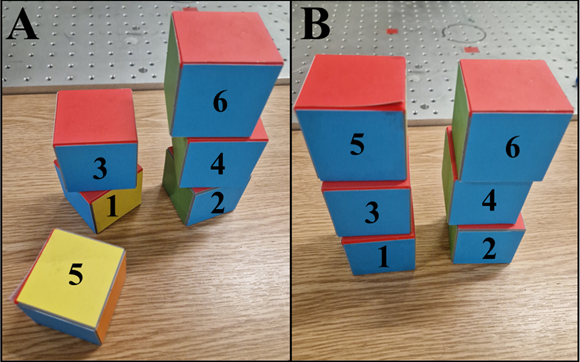}}
  \caption{Experimental stimuli}
  \label{fig:stimuli} 
\end{figure}

\begin{figure}[t]
    \centering
    \includegraphics[width=\columnwidth]{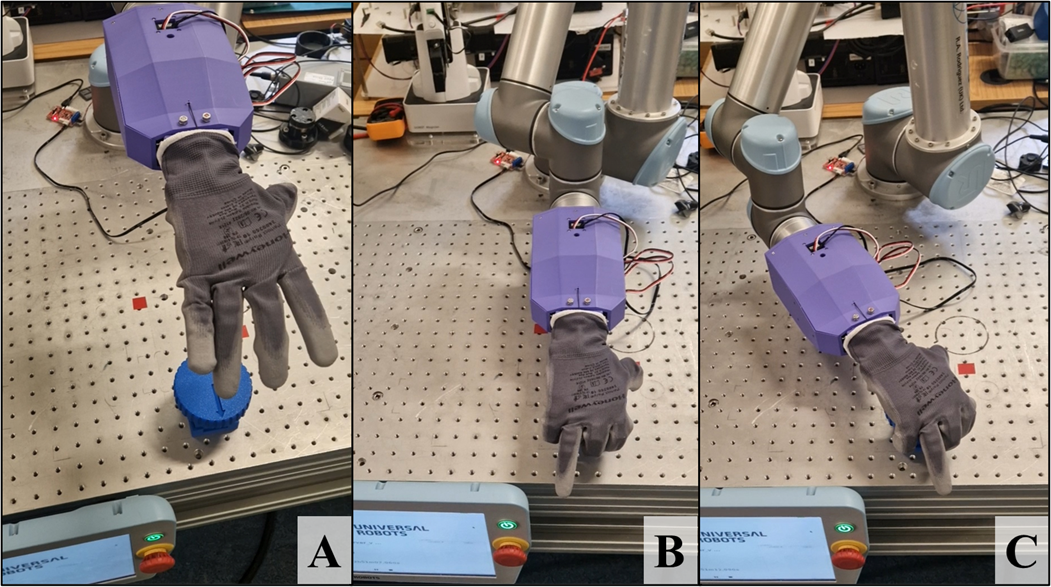}
    \caption{Snapshots of the rotation test under the no wrist condition. The hand lowers onto the disc (A), before grasping (B) and rotating (C). At this point the arm cannot rotate the disc any farther, so releases the disc and returns to position B, before grasping the disc and rotating to C again.}
    \label{fig:rotation_pics}
    \vspace{-1em}
\end{figure}

\begin{figure*}[t!]
    \centering
  \subfloat[\textbf{Without} actuation of the wrist.\label{fig:rot_no_wris}]{%
       \includegraphics[width=0.48\linewidth]{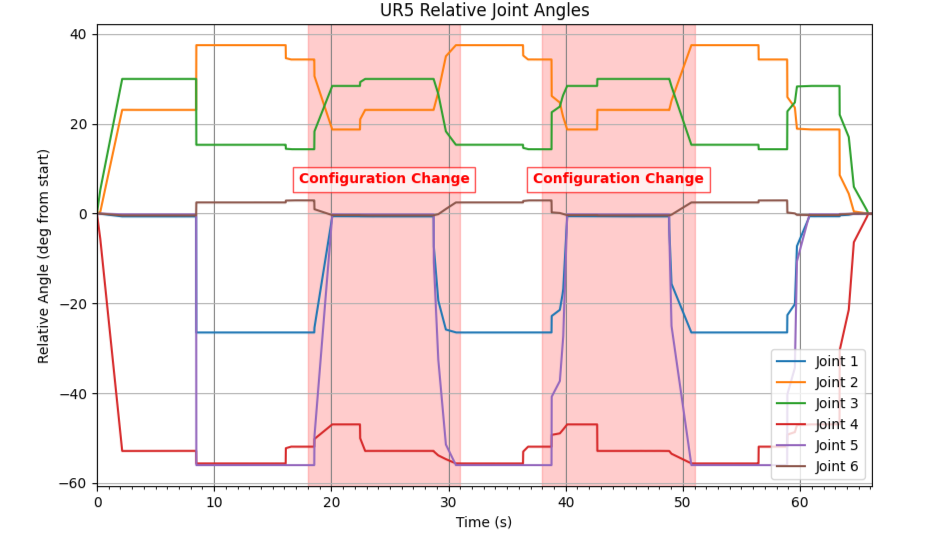}}
    \hfill
  \subfloat[\textbf{With} actuation of the wrist.\label{fig:rot_wrist}]{%
        \includegraphics[width=0.48\linewidth]{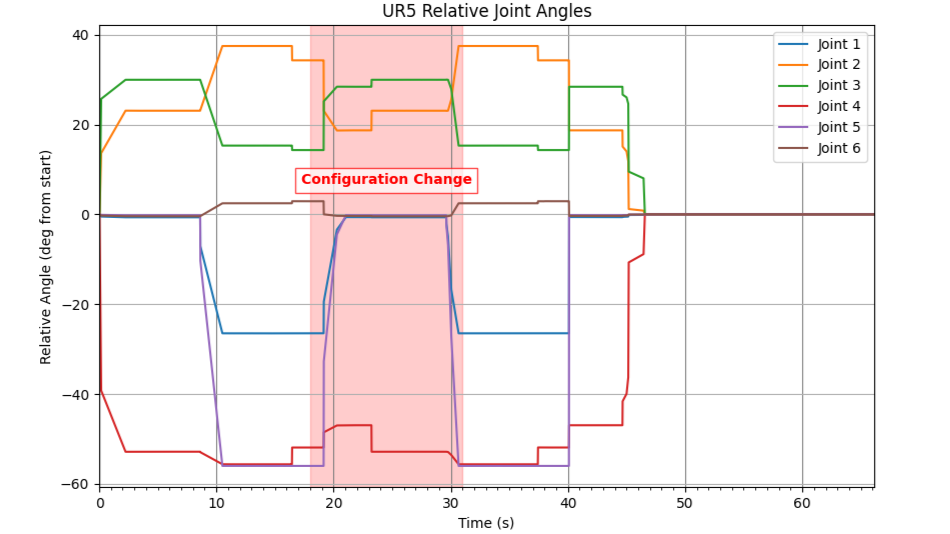}}
  \caption{Robot arm joint angles during the \textbf{object rotation} task, relative to initial positions. The configuration changes occur when the arm can not rotate the end-effector any farther, so must release the disc, untwist and grasp again. Wrist actuation results in a faster time to complete the task as only one configuration change is necessary (see also Fig.~\ref{fig:rotation_pics}).}
  \vspace{-1em}
  \label{fig:rotation} 
\end{figure*}

\subsection{Experimental Methodology}\label{sec:experiment_method}
For testing, the SoftWrist is mounted on a UR5 (Universal Robots) robotic arm: a 6-DoF industrial arm with reach 850\,\unit{mm} and payload 5\,\unit{kg}. Its final DoF is a rotational joint used for pronation/supination. It is controlled using a python library based on the api supplied by Universal Robotics. The arm is mounted on a table, with a workspace of 100\,\unit{mm} (width) by 130\,\unit{mm} (depth) in front of the arm.

\subsubsection{Rotation task}
A common action that utilises our wrists is screwing/unscrewing objects such as lids. This rotation task involves rotating a disc 90\unit{\degree} clockwise (Fig. 5A). The disk is 20\,\unit{mm} thick with diameter 90\,\unit{mm}, and is mounted on a fixed base with the hand is positioned above (Fig.~\ref{fig:rotation_pics}A). The hand is then commanded to grasp the disc (Fig.~\ref{fig:rotation_pics}B) and rotate it until either the arm requires a configuration change (Fig.~\ref{fig:rotation_pics}C), or the disc has been fully rotated through 90\unit{\degree}. The test is repeated under two conditions: with the wrist actuated and without. The goal is to compare the performance of the hand with and without the wrist in terms of time taken to complete the task and number of configuration changes required. For this test, a work glove was used to increase friction between the hand and disc (see Fig.~\ref{fig:rotation_pics}).

\subsubsection{Stacking task}
To test the reorientation capabilities of the hand-wrist combination, a stacking task is performed. Six 50\,\unit{mm} cubes six different-colored faces are placed on the table such that a distinct 90\unit{\degree} rotation is required for each cube to reach the same orientation (Fig.~\ref{fig:stimuli}). The task is to build a 3 layer structure with the cubes. The first two cubes are to be rotated $\pm$90\unit{\degree} in the yaw axis and placed next to each other. The next two cubes are to be rotated $\pm$90\unit{\degree} in the roll axis and placed on top of the first two cubes. The final two cubes are to be rotated $\pm$90\unit{\degree} in the pitch axis and placed on top of the last two cubes. The task is repeated with and without wrist actuation, to compare the performance of the hand in terms of time taken to complete the task, joint displacements, end effector path, and success rate. The task is considered successfully completed if all cubes are correctly stacked in the correct orientation.

\section{RESULTS}\label{Results}

\begin{figure}[b!]
    \vspace{-1em}
    \centering
    \includegraphics[width=0.75\columnwidth,trim={0 0 0 30},clip]{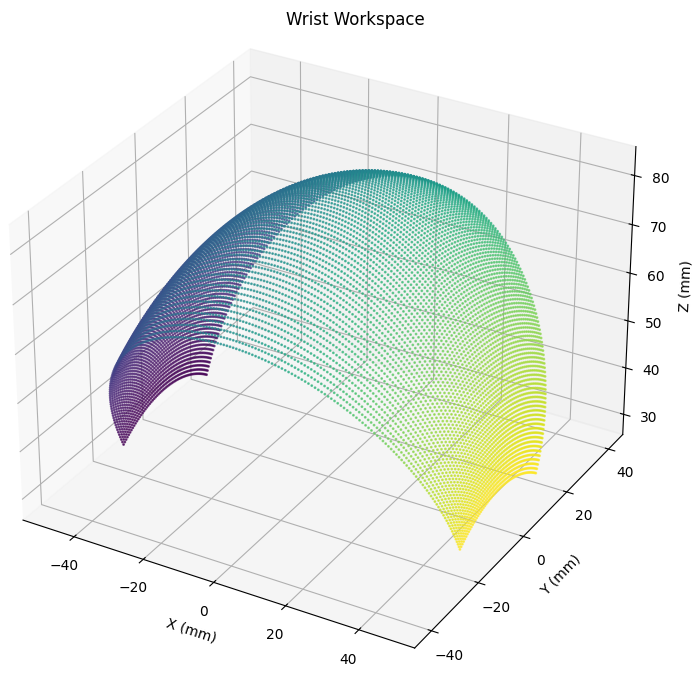}
    \caption{The reachable workspace of the centre of the palm, when the base of the wrist is fixed.}
    \label{fig:wrist_workspace}
\end{figure}

\subsection{General Capabilities}
Overall, the wrist is capable of 30\unit{\degree} ulnar and radial deviation (see Fig.~\ref{fig:main_photo}). The wrist is also capable of 90\unit{\degree} flexion and extension (Fig.~\ref{fig:main_photo}). This range of motion is also demonstrated in Fig.~\ref{fig:wrist_workspace}, which shows the workspace of the centre of the palm when actuating the wrist. Ryu \textit{et al.} found that the ideal range of motion for wrist flexion/extension and ulnar/radial deviation is 54\unit{\degree}, 60\unit{\degree}, 40\unit{\degree}, 17\unit{\degree}, respectively \cite{ryu1991functional}. It is also evident from Fig.~\ref{fig:main_photo}C,D that the fingers and wrist are successfully decoupled. Otherwise, when the fingers are completely open, coupling would cause the fingers to close when extending the wrist.


\begin{figure}[t!]
    \centering
    \vspace{-1em}
    \includegraphics[width=0.5\textwidth]{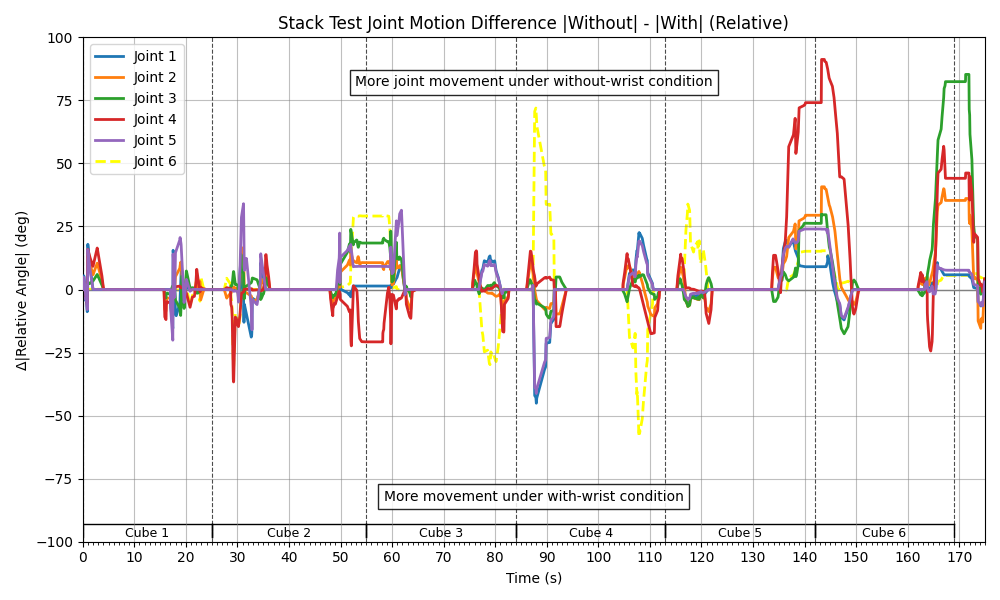}
    \vspace{-2em}
    \caption{Results from the \textbf{object stacking} task. Difference in absolute joint angles of the UR5 relative to the initial positions, between the fixed wrist and actuated wrist conditions. Positive values mean a joint moved more without wrist actuation. Negative values mean it moved more with wrist actuation. Note that large values in joint 6 (shown in yellow) are not important for this comparison, as it relates to rotation of the end effector and does not contribute to awkward arm configurations.}
    \label{fig:stack_joint}
\end{figure}

\subsection{Manipulation Tasks}

\begin{figure*}[t]
    \centering
  \subfloat[\textbf{Without} actuation of the wrist.\label{fig:stack_end_no_wrist}]{%
       \includegraphics[width=0.4\linewidth,trim={0 0 0 50},clip]{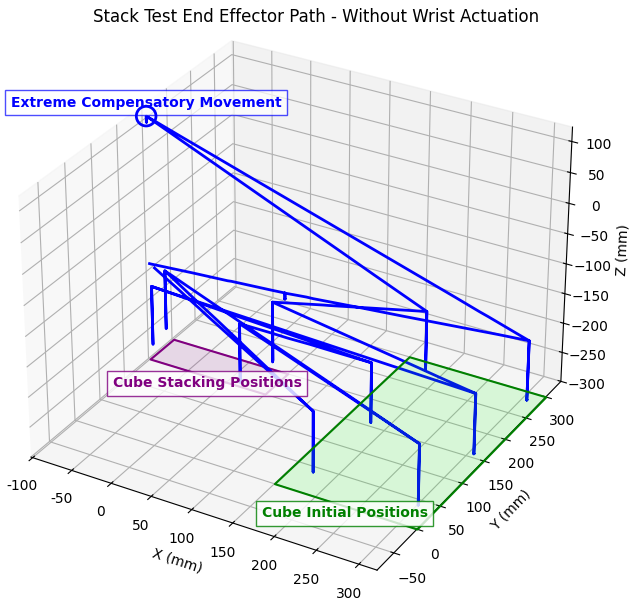}}
    \hfill
  \subfloat[\textbf{With} actuation of the wrist.\label{fig:stack_end_wrist}]{%
        \includegraphics[width=0.4\linewidth,trim={0 0 0 50},clip]{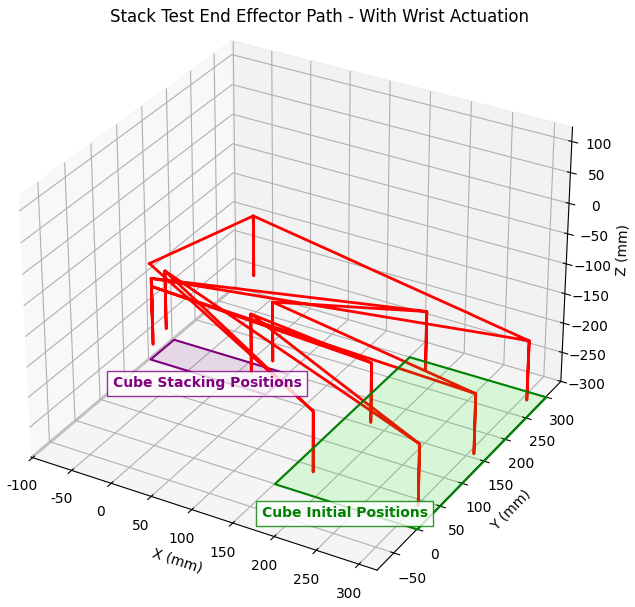}}
  \caption{Pose of the end-effector during the stacking test. Cubes are picked from the green area and stacked in the purple area. Without wrist actuation, an extreme compensatory movement is necessary for the arm to pitch the cube upward, resulting in a failed placement of cube 5.}
  \vspace{-1em}
  \label{fig:stack_end} 
\end{figure*}

\subsubsection{Object Rotation Task}
The hand was able to rotate the disc 90\unit{\degree} in both conditions, with the wrist actuated and without~(Figs~\ref{fig:rotation_pics}). However, the time taken to complete the task was significantly reduced when the wrist was actuated~(Fig.~\ref{fig:rotation}). A reconfiguration change is needed part way through the movement when the arm can not rotate the end-effector any farther, so must release the disc, untwist and grasp again. The time taken to complete the task with the wrist actuated was 47\,\unit{s}, compared to 66\,\unit{s} without the wrist, resulting in a $\sim$29\% reduction in completion time. The number of configuration changes required was also reduced when the wrist was actuated, with only 1 configuration change required compared to 2 without the wrist. These configuration changes were necessary to prevent the arm from operating past a joint limit, entering a singularity, or colliding with itself or the bounds of the work area. 


\begin{table}[t!]
\centering
\small
\setlength{\tabcolsep}{4pt}
\begin{tabular}{l cc}
\toprule
\textbf{Task / Capability} & \textbf{No Wrist} & \textbf{With Wrist} \\
\midrule
General Range & \multicolumn{2}{c}{$\pm30^\circ$ U/R Dev, $\pm90^\circ$ F/E} \\
\midrule
\textbf{Rotation Task} & & \\
\quad Config Changes & 2 & \textbf{1} \\
\quad Completion Time & 66s & \textbf{47s} \\
\midrule
\textbf{Stacking Task} & & \\
\quad Success (Stack) & 5/6 & \textbf{6/6} \\
\quad Success (Reorient) & 4/6 & \textbf{6/6} \\
\bottomrule
\end{tabular}
\caption{Summary of SoftHand-W capabilities and results.}
\vspace{-3em}
\label{tab:results}
\end{table}

\subsubsection{Cube Stacking Task}
When attempting the stacking test without actuating the wrist, the arm managed to correctly reorient 4 of the 6 cubes and successfully stack 5 (end result shown in Fig.~\ref{fig:stimuli}A). Only a partial rotation of cube 1 was achieved to prevent the arm from colliding with the table. The arm also had to make a very large compensatory movement to place the 5th cube, as shown in Fig.~\ref{fig:stack_end_no_wrist}. Without use of the wrist, the only way to pitch the cube upwards was to move the arm in a large arc, raising the cube significantly above the stack before letting go, resulting in a failed placement. This is evident in the joint angles (Fig.~\ref{fig:stack_joint}, upper half of plot), which also shows a large amount of joint movement for cubes 5 and 6, particularly in joints 3 and 4.

When actuating the wrist, the hand was able to successfully reorient and stack all 6 cubes (end results shown in Fig.~\ref{fig:stimuli}B). The wrist allowed the hand to rotate the first cube fully and pitch the 5th cube upwards without lifting the cube vertically, as shown in Fig.~\ref{fig:stack_end_wrist}. This was impossible without the wrist, demonstrating the benefit to dexterity of the additional DoFs. The joint angles (Fig.~\ref{fig:stack_joint}), show a significant reduction in movement, particularly in joints 3 and 4, which are responsible for pitch and yaw of the end-effector. Joint 4 shows the largest difference, with roughly 36\% lower maximum joint travel when using the wrist.

\section{DISCUSSION}\label{Discussion}
In this paper, we presented the SoftHand-W: a 3D-printed, underactuated, tendon-driven anthropomorphic hand with an actively actuated 2 DoF wrist and biomimetic carpal tunnel tendon routing. Using just a four-servo control system, the hand achieves self-adaptive grasping and gains additional object reorientation capability from the wrist, which greatly reduces the burden on the robot arm during manipulation tasks. In particular, the wrist reduces the need for large compensatory arm motion during manipulation. Experiments showed improved task efficiency with wrist actuation improving success rate in the stacking task by 50\% and reducing task completion time by 29\% in the rotation task (results summarized in Table~\ref{tab:results}), indicating that improved wrist dexterity can materially aid manipulation while preserving a compact, printable design and simple control.

Performance was good for grasping light, medium-sized objects and reorientation tasks. Finger-wrist decoupling via PTFE sheaths prevented observable involuntary finger motion during extreme wrist poses, as is evident in the Figure~\ref{fig:main_photo}. The wrist's range of motion is close to human limits, and limited by the positioning of the deviation cams and the M4 bolts that fasten the wrist to the actuation box. Repositioning these bolts would allow up to 45\unit{\degree} in both directions, which would allow the SoftHand-W to exceed the ideal (human) range of motion in all directions. 


Other improvements could be made to the design to further enhance the hand's dexterity: introducing an adjustable (or actuated) thumb base angle to improve opposition, adopting individual and adjustable tendon terminations such as turnbuckles for fine tension trim, and replacing the plain PTFE tubes with reinforced Bowden tubes to prevent bunching or collapse. Additional future work could include finger splay and an added thumb DoF to expand the grasp set, integration of fingertip tactile sensing for force feedback, and faster servos to enable dynamic manoeuvrers like throwing.
                                  
Overall, the SoftHand-W shows that adding a wrist meaningfully enhances reorientation performance for an underactuated anthropomorphic robot hand, advancing accessible pathways toward more human-like dexterous manipulation. The design's novel integration of active wrist articulation, synergy-driven underactuation, and carpal tunnel tendon routing provides a novel, reproducible foundation and benchmark for accessible dexterous robotics.





\bibliographystyle{IEEEtran}
\bibliography{references}

\end{document}